\documentclass[a4paper]{llncs}
\usepackage{amssymb}
\usepackage{graphicx}
\usepackage{rotating,textcomp}
\usepackage{url}
   
\newcommand{\keywords}[1]{\par\addvspace\baselineskip
\noindent\keywordname\enspace\ignorespaces#1}

\begin{document}

\mainmatter  

\title{New Mechanism of Combination Crossover Operators in Genetic Algorithm for Solving the Traveling Salesman Problem}

\author{Pham Dinh Thanh \inst{1} \and Huynh Thi Thanh Binh \inst{2} \and Bui Thu Lam \inst{3}}

\institute{Tay Bac University  \and Hanoi University of Science and Technology \and Le Quy Don Technical University}

\maketitle

\begin{abstract}
Traveling salesman problem (\textit{TSP}) is a well-known in computing field. There are many researches to improve the genetic algorithm for solving \textit{TSP}. In this paper, we propose two new crossover operators and new mechanism of combination crossover operators in genetic algorithm for solving \textit{TSP}. We experimented on \textit{TSP} instances from \textit{TSP}-Lib and compared the results of proposed algorithm with genetic algorithm (\textit{GA}), which used \textit{MSCX}. Experimental results show that, our proposed algorithm is better than the \textit{GA} using \textit{MSCX} on the min, mean cost values.
\end{abstract}

\section{Introduction}
\indent The traveling salesman problem is an important problem in computing fields and has many applications in the daily life such as scheduling, vehicle routing, \textit{VLSI} layout design, etc. The problem was first formulated in 1930 and it has been one of the most intensively studied problems in optimization techniques. Until now, researchers have obtained numerous significant results for this problem.

\textit{TSP} is defined as following: Let 1, 2, \ldots, \textit{n} is the labels of the \textit{n} cities and $C = [c_{i,j}]$ be an $n \times n$ cost matrix where $c_{i,j}$ denotes the cost of traveling from city $i$ to city $j$. \textit{TSP} is the problem of finding the \textit{n}-city closed tour having the minimum cost such that each city is visited exactly once. The total cost A of a tour is.
\begin{equation}
A(n)=\sum_{i=1}^{n-1} c_{i,i+1} + c_{n,1}
\end{equation}
\textit{TSP} is formulated as finding a permutation of \textit{n} cities, which has the minimum cost. This problem is known to be \textit{NP}-hard \cite{ref:pp-01,ref:pp-02} but it can be applied in many real world applications \cite{ref:pp-13} so a good solution would be useful. 

Many algorithms have been suggested for solving \textit{TSP}. \textit{GA} is an approximate algorithm based on natural evolution, which applied to many different types of the combinatorial optimization. \textit{GA} can be used to find approximate solutions for \textit{TSP}.

There  are  a  lot  of  improvements in \textit{GA} that have been developed to increase the performance in solving the \textit{TSP} such as: optimizing creating initial population \cite{ref:pp-03}, improving mutation operator \cite{ref:pp-17}, creating new crossover operator  \cite{ref:pp-12,ref:pp-20,ref:pp-21,ref:pp-22,ref:pp-23,ref:pp-24,ref:pp-25}, combining with local search  \cite{ref:pp-04,ref:pp-06,ref:pp-07,ref:pp-08,ref:pp-18}.

In this paper, we introduce two new crossover operators: \textit{MSCX\_Radius} and \textit{RX}. We propose new mechanism of combination proposed crossover operators and \textit{MSCX} \cite{ref:pp-25} in \textit{GA} to solve \textit{TSP}. This combination is expected to adapt the changing of population. We experimented on \textit{TSP} instances from \textit{TSP}-Lib and compared the results of proposed algorithm with \textit{GA} which used \textit{MSCX}. Experimental results show that, our proposed algorithm is better than the \textit{GA} using \textit{MSCX} on the min, mean cost values.

The rest of this paper is organized as follows. In section 2, we will present related works. Section 3 and 4 contain the description of our new crossovers and the proposed algorithm for solving \textit{TSP} respectively. The details of our experiments and the computational and comparative results are given in section 5. The paper concludes with section 6 with some discussions on the future extension of this work.

\section{Related work}
\textit{TSP} is \textit{NP}-hard problems. There are two approaches for solving \textit{TSP}: exact and approximate. Exact approaches are based on Dynamic Programming \cite{ref:pp-14}, Branch and Bound \cite{ref:pp-02}, Integer Linear Programming \cite{ref:pp-21}, etc. Exact approaches used to give the optimal solutions for \textit{TSP}. However, these algorithms have exponential running time, therefore they only solved small instances. As M. Held and R. M. Karp \cite{ref:pp-14} pointed out Dynamic Programming takes $O(n^2\cdot2^n)$ running time, so that it only solves \textit{TSP} with a small number of the vertices.

In recent years, approximation approaches for solving \textit{TSP} are interested by researchers. These approaches can solve large instances and give approximate solutions near to the optimal solution (sometime optimal). Approximation approaches for solving \textit{TSP} are 2-opt, 3-opt \cite{ref:pp-01}, simulated annealing \cite{ref:pp-07,ref:pp-16}, tabu search  \cite{ref:pp-07,ref:pp-16}; nature based optimization algorithms and population based optimization algorithms: genetic algorithm \cite{ref:pp-03,ref:pp-06,ref:pp-07,ref:pp-08,ref:pp-10,ref:pp-11,ref:pp-12,ref:pp-13,ref:pp-16,ref:pp-17,ref:pp-19,ref:pp-22,ref:pp-25}, neural networks \cite{ref:pp-15}; swarm optimization algorithms: ant colony optimization \cite{ref:pp-07,ref:pp-23}, bee colony optimization \cite{ref:pp-18}.

\textit{GA} is one of computational model inspired by evolution, which has been applied to a large number of real world problems. \textit{GA} can be used to get approximate solutions for \textit{TSP}. High adaptability and the generalizing feature of \textit{GA} help to execute the traveling salesman problem by a simple structure.

M. Yagiura and T. Ibaraki \cite{ref:pp-16} proposed \textit{GA} for three permutation problems including \textit{TSP}; and \textit{GA} solving \textit{TSP} uses \textit{DP} in its crossover operator. The experiments are executed on 15 randomized Euclidean instances (5 instances for each \textit{n} = 100, 200, 500). The proposed algorithm \cite{ref:pp-16} could get better solutions than Multi-Local, Genetic-Local and Or-opt when sufficient computational time was allowed. However, the experimental results have been pointed out that, their proposed algorithm is ineffective to compare some heuristics specially designed to the given \textit{TSP}, such as Lin-Kernighan method \cite{ref:pp-01,ref:pp-16}.

In \cite{ref:pp-05}, the authors used local search and \textit{GA} for solving \textit{TSP}. The experiments are executed in kroA100, kroB100 and kroC100 instances. The experiments results show that the combination of two genetic operators, \textit{IVM} and \textit{POS}, and 2-opt have better cost for solving \textit{TSP} problem. However, the algorithm took more time to converge to the global optimum than using 3-opt.

In 1997, Bernd Freisleben, Peter Merz \cite{ref:pp-07} proposed Genetic Local Search for the \textit{TSP}. This algorithm used idea of hill climber to develop local search in \textit{GA}. The experiment shows that the best solutions are better than the one in \cite{ref:pp-24} on running time and better on min cost range from 0.46\% $\to$ 0.21\%.

Crossover operator is one of the most important component in \textit{GA}, which generates new individual(s) by combining genetic material from two parents but preserving gene from the parents. The researchers have studied many different optimal crossover operators like creating new crossover operators \cite{ref:pp-22,ref:pp-24}, modifying exist crossover operators \cite{ref:pp-20,ref:pp-21,ref:pp-23,ref:pp-25}, and hybridizing crossover operators \cite{ref:pp-10}.

Sehrawat, M. et al. \cite{ref:pp-20} modified Order Crossover (\textit{OX}). They selected the first crossover point which is the first node of the minimum edge from second chromosome. The experiment was executed on five sample data. The modifying order crossover (\textit{MOX}) could get better solutions than \textit{OX} on two sample data but number of the best solutions is found by \textit{MOX} more than \textit{OX}.

The new genetic algorithm (called \textit{FRAG\_GA}) was developed by Shubhra, Sanghamitra and Sankar \cite{ref:pp-21}. There were two new operators: nearest fragment (\textit{NF}) and modified order crossover (\textit{MOC}). The \textit{NF} is used for optimizing initial population. In the \textit{MOC}, the authors performed two changes: length of a substring for performing order crossover is $y = \max\{2,\alpha\}$, where $n/9 \leq  \alpha \leq n/7$ (\textit{n} is the total number of cities) and the length of substring is predefined at any times performing crossover. The experiments are executed in Grtschels24, kroA100, d198, ts225, pcb442 and rat783 instances. The authors compared \textit{FRAG\_GA} with \textit{SWAPGATSP} \cite{ref:pp-12} and \textit{OXSIM} (standard \textit{GA} with order crossover and simple inversion mutation) \cite{ref:pp-13}. The experiment results showed that the best result, the average result and computation time of \textit{FRAG\_GA} are better than one of \textit{SWAPGATSP}, \textit{OXSIM}.

In \cite{ref:pp-22}, the authors proposed an improving \textit{GA} (\textit{IGA}) with a new crossover operator (Swapped Inverted Crossover - \textit{SIC}) and a new operation called Rearrangement. \textit{SIC} creates 12 children from 2 parents then select 10 for applying multi mutation. Finally select 2 best individuals. Rearrangement Operation is applied to all individuals in population. It finds the maximum cost of two adjacent cities then swap one city with three other cities. The experiments are executed 10 times for each instances (KroA100, D198, Pcb442 and Rat783). The experiments show that, performance of \textit{IGA} is better than the three compared \textit{GAs}.

Kusum and Hadush \cite{ref:pp-23} modified the \textit{OX}. In these proposing crossovers, the positions of cut points or the length of the substrings in both parents are different. The experimented on six Euclidean instances derived from \textit{TSP}-lib (eil51, eil76, kroA100, eil101, lin105 and rat195). Crossover rate is 0.9 and mutation rate is 0.01. The experimental results show that results of one modifying crossover are better than \textit{OX} for six \textit{TSP} instances.

In \cite{ref:pp-24}, the authors proposed new crossover operator, Sequential Constructive crossover (\textit{SCX}). The main idea of \textit{SCX} is selecting the edges having less value based on maintaining the sequence of cities in the parents. The experiments are performed in 27 \textit{TSPLIB} instances. Results of experiment show that \textit{SCX} is better than the \textit{ERX} and \textit{GNX} on quality of solutions and solution times.

In 2012, Sabry, Abdel-Moetty and Asmaa \cite{ref:pp-25} proposed new crossover operator, Modified Sequential Constructive crossover (\textit{MSCX}), which is an improvement of the \textit{SCX} \cite{ref:pp-24}. The \textit{MSCX} create an offspring and description as follows:\\
\indent \textbf{Step 1}: Start from \textquotesingle First Node\textquotesingle \ of the parent 1 (i.e., current node p = parent1(1)).\\
\indent \textbf{Step 2}:  Sequentially search both of the parent chromosomes and consider\\
\indent The first \textquotesingle legitimate node\textquotesingle \ (the node that is not yet visited) appeared after \textquotesingle node p\textquotesingle \ in each parent. If no \textquotesingle legitimate node\textquotesingle \ after node p is present in any of the parent, search sequentially the nodes from parent 1 and parent 2 (the first \textquotesingle legitimate node\textquotesingle \ that is not yet visited from parent1 and parent2), and go to Step 3.\\
\indent \textbf{Step 3}: Suppose the \textquotesingle Node $\alpha$\textquotesingle \ and the \textquotesingle Node $\beta$\textquotesingle \ are found in 1st and 2nd parent respectively, then for selecting the next node go to Step 4.\\
\indent \textbf{Step 4}: If $C_{p\alpha}< C_{p\beta}$, then select \textquotesingle Node $\alpha$\textquotesingle , otherwise, \textquotesingle Node $\beta$\textquotesingle \ as the next node and concatenate it to the partially constructed offspring chromosome. If the offspring is a complete chromosome, then stop, otherwise, rename the present node as \textquotesingle Node p\textquotesingle \  and go to Step 2.

Although a lot of crossovers were developed for solving \textit{TSP}, but each operator has its property, so, in this paper, we propose two new crossover operators and mechanism of combination them with \textit{MSCX} crossover \cite{ref:pp-25}. This scheme is expected to adapt the changing and convergence of population and improve the effectiveness in terms of cost of tour. The proposed algorithm will be presented in the next section.

\section{Proposed crossover operators}
This section introduces two new crossover operators: \textit{MSCX\_Radius}, \textit{RX}, which are developed for improving the best solutions and increase the diversity of the population. 

\subsection{MSCX\_Radius Crossover}
\textit{MSCX\_Radius} modify the step two of \textit{MSCX} \cite{ref:pp-25}. In \textit{MSCX\_Radius}, if no \textquotesingle legitimate node\textquotesingle \ after current node, find sequentially r nodes, which are not visited from the parents. Then select the node having the smallest distance to current node. r is parameter of \textit{MSCX\_Radius}.

\subsection{RX Crossover}
This crossover operator is described as following:\\
\indent \textbf{Step 1}: Randomly select \textit{pr}\% cities from the first parent to the offspring.\\
\indent \textbf{Step 2}: Copy the remaining unused cities into the offspring in the order they appear in the second parent.\\
\indent \textbf{Step 3}: Create the second offspring in an analogous manner, with the parent roles reverse.

\begin{figure}[!htp]
	\centering
	\includegraphics[scale=0.7]{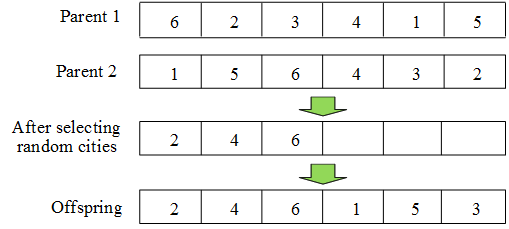}
	\caption{Illustration of the RX crossover operator, pr\% = 20\%}
	\label{fig:rx_cross_exam}
\end{figure}

Figure \ref{fig:rx_cross_exam} show an example of \textit{RX} crossover operator.

\section{Proposed mechanism of combination two new crossovers and MSCX}
This section proposes new mechanism of combination two propose crossover operators \textit{MSCX\_Radius} and \textit{RX} with \textit{MSCX}. We then use apply this mechanism in an improving genetic algorithm (\textit{CXGA}) for solving \textit{TSP}.

\begin{figure}[!htp]
	\centering
	\includegraphics[scale=0.5]{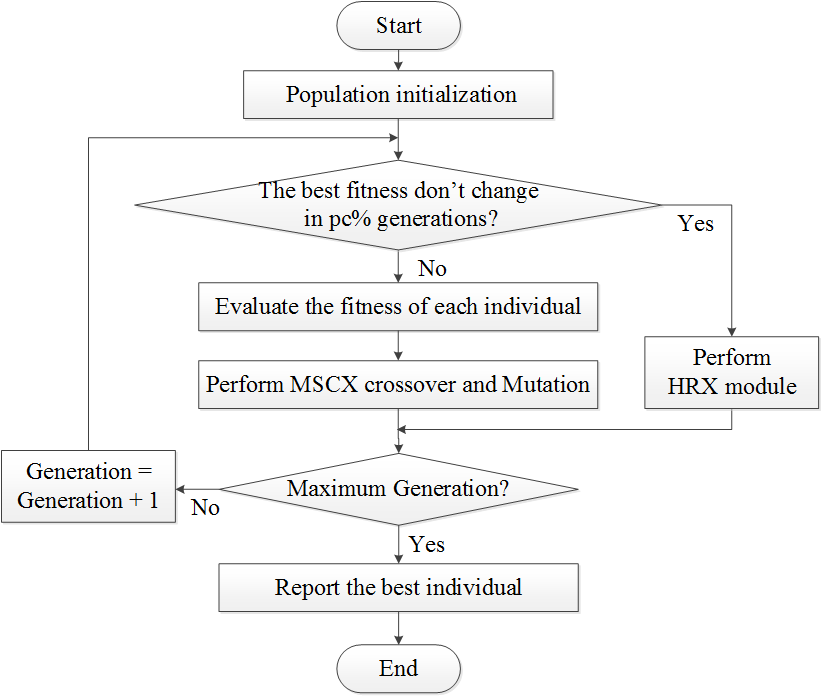}
	\caption{Structure of improved genetic algorithm}
	\label{fig:IGA_Structure}
\end{figure}

The workflow of \textit{CXGA} is described in Fig.\ref{fig:IGA_Structure}.

\begin{figure}[!htp]
	\centering
	\includegraphics[scale=0.45]{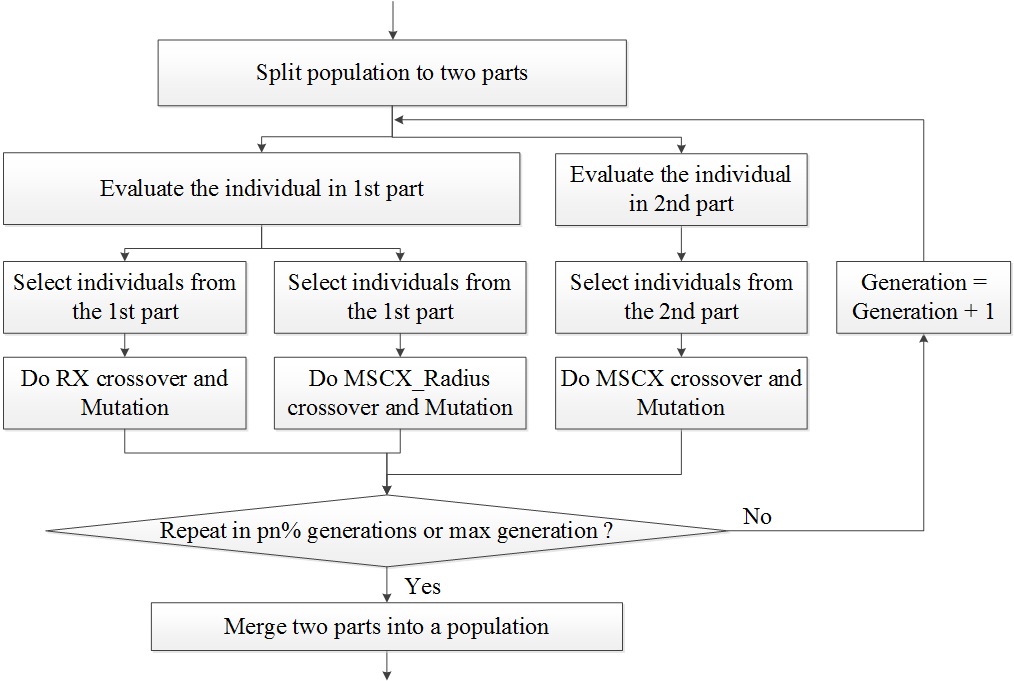}
	\caption{Structure of HRX module in CXGA}
	\label{fig:HRX_Module_Structure}
\end{figure}

The workflow of \textit{HRX} module is shown in Fig.\ref{fig:HRX_Module_Structure}.

In the first part, \textit{prx}\% of individuals will be chosen for \textit{R}X crossover and the rest for \textit{MSCX\_Radius}

Sketch of the \textit{HRX} module is presented as below:

\begin{verbatim}
Procedure: HRX (P, prx, pr, r)
Input: 	The population P
        r: parameter of MSCX_Radius
        prx: percent of individuals from first part use RX
        pr: percent of number of cities is gotten random in RX
Output: The optimization population P'
\end{verbatim}
\verb|Begin|\\
\verb|  Split P into two parts: P1 and P2;|\\
\verb|  i| $\gets$ \verb|0;| $\mbox{FP}_i \gets$ \verb|P2;| $ \mbox{SP}_i$ $\gets$ \verb|P1; numInRX| $\gets$ \verb+(prx * |P1|)/100;+ \\
\verb|  ng| $\gets$ \verb|number of generations perform HRX module;|\\
\verb|  While i < ng do| \\
\verb|     For j := 1 to numInRX/2 do|\\
\verb|       Select random individuals from| $\mbox{SP}_i$\verb|;|\\
\verb|       Do RX(pr)crossover, mutation;|\\
\verb|       Add offsprings to| $\mbox{SP}_{i+1}$\verb|;|\\
\verb|     End for|\\
\verb+     For j :=1 to |P1| - numInRX do+\\
\verb|       Select random individuals from| $\mbox{SP}_i$\verb|;|\\
\verb|       Do MSCX_Radius(r) crossover, mutation;|\\
\verb|       Add offspring to| $\mbox{SP}_{i+1}$;\\
\verb|     End for|\\
\verb+     For j := 1 to |P2| do+\\
\verb|       Select random individuals from |$\mbox{FP}_i$ \verb|;|\\
\verb|       Do MSCX crossover, mutation; |\\
\verb|       Add offspring to| $\mbox{FP}_{i+1}$\verb|;|\\
\verb|     End for|\\
\verb|     i| $\gets$ \verb|i + 1;|\\
\verb|  End while|\\
\verb|  Merge FPi, SPi into P';|\\
\verb|  Return P'|\\
\verb|End;|
\section{Computational results}
\subsection{Problem instances}
The results are reported for the symmetric \textit{TSP} by extracting benchmark instances from the \textit{TSP}-Lib \cite{ref:url-09}. The instances chosen for our experiments are eil51.tsp, Pr76.tsp, Rat99.tsp, KroA100, Lin105.tsp, Bier127.tsp, Ts225.tsp, \linebreak Gil262.tsp, A280.tsp, Lin318.tsp, Pr439.tsp and Rat575.tsp. The number of vertices: 51, 76, 99, 100, 105, 127, 225, 262, 280, 318, 439, 575. Their weights are Euclidean distance in 2-D.

\subsection{System setting}
In the experiment, the system was run 10 times for each problem instance. All the programs were run on a machine with Intel Pentium Duo E2180 2.0GHz, 1GB RAM, and were installed by C\# language.

\subsection{Experimental setup}
This paper implemented two sets of experiments. In the first, we run \textit{GA} using \textit{MSCX\_Radius} (named \textit{GA1}), \textit{GA} using \textit{RX} (named \textit{GA2}) and compare with \textit{GA} using \textit{MSCX} \cite{ref:pp-25} (named \textit{GA3}). In the second, we compare the performance of \textit{CXGA} with \textit{GA3}. 

When execute the \textit{HRX} module, the population is split into two part. The first one includes the best solutions which uses a combination of \textit{MSCX\_Radius} and \textit{RX} crossover; the second includes the rest solutions of population, which uses \textit{MSCX} crossover.

The parameters for experiments are:\\
\indent \indent Population size: $p_s = 100$\\
\indent \indent Number of evaluation: 1000000\\
\indent \indent  Mutation rate: $p_m$ = 1/number\_of\_city (chromo length)\\
\indent \indent Crossover rate: $p_c = 0.9$

\subsection{Experimental resultstitle}
The experiments were implemented in order to compare \textit{GA1}, \textit{GA2}, \textit{CXGA} with \textit{GA3} in term of the min, mean, standard deviation values and running times.

For comparing effects of two new crossover operations: \textit{MSCX\_Radius} and \textit{RX}. We tested \textit{GA1}, \textit{GA2} with different values of \textit{r}, \textit{pr} parameters. The best results obtain by \textit{GA1}, \textit{GA2} are compared with the ones obtain by \textit{GA3}.

\begin{figure}
	\centering
	\includegraphics[scale=0.7]{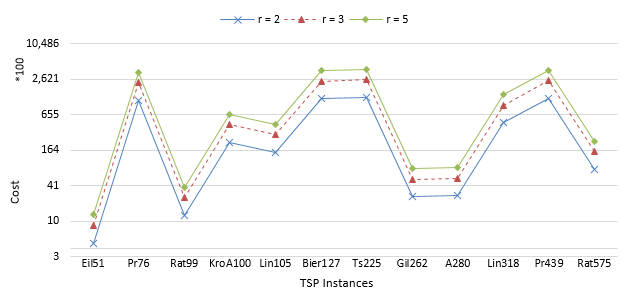}
	\caption{The mean cost on TSP instances of GA1 when r = 2, 3 and 5}
	\label{fig:results_change_r}
\end{figure}

Figure \ref{fig:results_change_r} summarizes the mean cost of \textit{GA1} when \textit{r} = 2, 3 and 5 respectively. With \textit{r} = 2, the results found by \textit{GA1} are the best.

\begin{figure}
	\centering
	\includegraphics[scale=0.6]{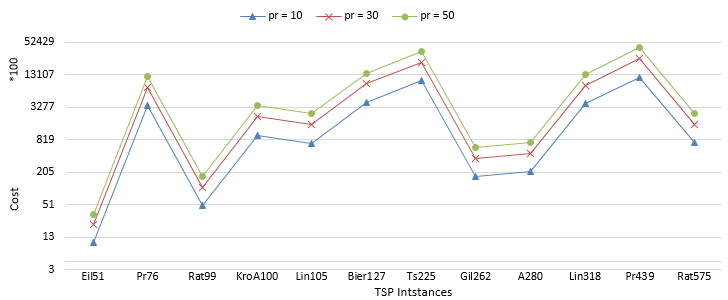}
	\caption{The mean cost on TSP instances of GA2 when pr\% = 10\%, 30\% and 50\%}
	\label{fig:results_change_pr}
\end{figure}

The Fig. \ref{fig:results_change_pr} illustrates the mean cost of \textit{GA2} when \textit{pr}\% = 10\%, 30\% and 50\% respectively. The diagrams show that, the mean cost of \textit{GA2} when \textit{pr}\% = 10\% are better than ones when \textit{pr}\% = 30\%, 50\%.

Experiment results on Fig. \ref{fig:results_change_r}, Fig. \ref{fig:results_change_pr} show that \textit{r} = 2 and\textit{ pr}\% = 10\% are the best parameters for \textit{GA1} and \textit{GA2} and they will be selected for comparison with \textit{GA3}.

\textit{HRX} module was implemented in differences parameters to find the best parameter. The size of the first part is 90\%, \textit{pc} =15\%, \textit{r} = 5, \textit{pr} = 30\%, \textit{pn} = 5\%, \textit{prx} = 40\%.

In order to select the best value of pc in \textit{HRX} module, we analyzed the correlative of the best solution obtaining from \textit{CXGA} with different values of \textit{pc} parameter (\textit{pc}\% = 5\%, 10\%, 15\%, 20\%, 30\% and 50\%).

\begin{figure}[!htp]
	\centering
	\includegraphics[scale=0.9]{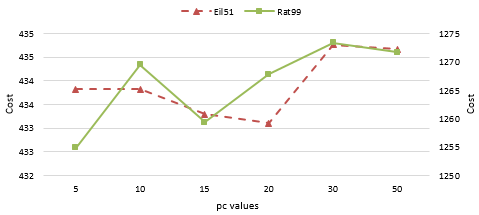}
	\caption{The relationship between the pc values, mean cost found by 10 running times of CXGA on Eil51, Rat99 instance}
	\label{fig:6_results_select_pc}
\end{figure}

The Fig. \ref{fig:6_results_select_pc} shows the dependence between the \textit{pc} values, mean cost values found on 10 running times of \textit{CXGA}. According to the experiments in the Fig. \ref{fig:6_results_select_pc}, \textit{pc}\% = 15\% is quite reasonable in our algorithm.

In \textit{MSCX\_Radius} crossover, the bigger the \textit{r} parameter is, the more increasing the running times is. In addition, according to the results in the Table \ref{tab:1-results of CXGA}, the results of \textit{CXGA} when \textit{r} = 5 are better than the ones when \textit{r} = 2, 3, 7 and 10 in most instances on mean and min values (values in bold). So, we chose 5 in all experiments for \textit{r} value.

Table \ref{tab:2-Compare results} summarizes the results found by \textit{GA3}, \textit{CXGA} and the best results of \textit{GA1}, \textit{GA2} for 12 \textit{TSP} instances of size from 51 to 575. 

Mean, min cost value found by \textit{GA1} are worse than \textit{GA3} on 8/12 and 7/12 instances. Standard deviation values found by \textit{GA1} worse than \textit{GA3} on 3 instances. The running time of \textit{GA1} are lower than \textit{GA3} on 3/12 instances. The running time of \textit{GA2} are faster than \textit{GA3} on all instances. Min, mean and standard deviation values found by \textit{GA2} are greater than \textit{GA3} about three times on all instances.

The mean cost values found by \textit{CXGA} algorithm are better than the ones found by \textit{GA3} from 0.2\% to 2.4\%. The min cost found by \textit{CXGA} are better than the one found by \textit{GA3} from 0.1\% to 2.8\%. The running time of \textit{CXGA} are faster than the ones found by \textit{GA3} on 11/12 instances. The standard deviation values found by \textit{CXGA} are better than \textit{GA3} on 7/12 instances (values in bold).

\section{Conclusion}
In this paper, we propose two new crossover operators, called \textit{RX} and \linebreak \textit{MSCX\_Radius}, and new mechanism of combination in \textit{GA} to adapt the convergence of the population for solving \textit{TSP}. We experimented on 12 Euclidean instances derived from \textit{TSP}-lib with the number of vertices from 51 to 575. Experiment results show that, the proposed combination crossover operators in \textit{GA} is effective for \textit{TSP}.

In the future, we are planning to apply propose mechanism of combination to another optimization problem.

\newpage
\begin{table}[htbp]
	\centering
\begin{sideways}
	\begin{minipage}{24cm}
		\caption{The results of CXGA found by 10 running times on Eil51, Pr76, Rat99, KroA100, Lin105, Bier127, Ts225, Gil262, A280 when r = 2,3,5,7,10}
		\includegraphics[scale=0.8]{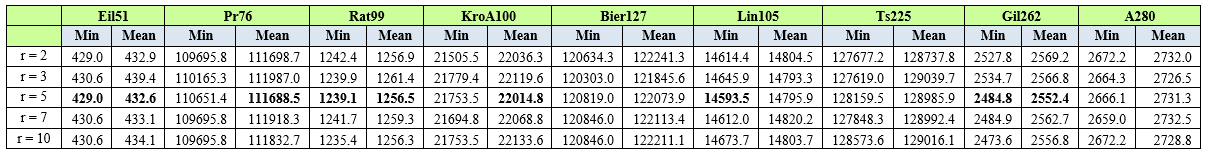}
		\label{tab:1-results of CXGA}
		\hfill {\scriptsize Min: Minimum cost;  Mean: Mean cost}
		
		\vspace{1cm}
		\caption{THE RESULTS FOUND BY GA1, GA2, GA3 AND CXGA ALGORITHM FOR TSP INTSTANCES}	
		\includegraphics[scale=0.8]{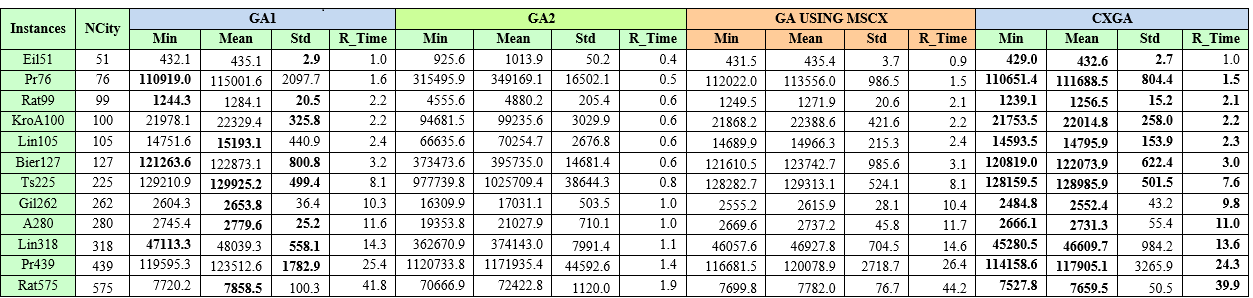}
		\label{tab:2-Compare results}
		\hfill {\scriptsize NCity: Number of city; Min: Minimum cost;  Mean: Mean cost; Std: Standard Deviation of minimum cost; R\_Time: Running time  (minutes)}	
	\end{minipage}
\end{sideways}
\end{table}


\begin{thebibliography}{4}

\bibitem{ref:pp-01} 1.	Lin, S., Kernighan, B.W.: An effective heuristic algorithm for the traveling salesman problem. Operations Research, vol. 21, pp. 498–-516 (1973)

\bibitem{ref:pp-02} Eiben, A.E., Smith, J.E.: Introduction to Evolutionary Computing Natural Computing. Series 1st edition. Springer (2003)

\bibitem{ref:pp-03} Snyder, L.V., Daskin, M.S.: A random-key genetic algorithm for the generalized traveling salesman problem. European Journal of Operational research, vol. 174, pp. 38–-53 (2006)

\bibitem{ref:pp-04} Paquete, L., Stützle, T.: A Two-Phase Local Search for the Biobjective Traveling Salesman Problem. In: Second Int. Conf. (EMO 2003), pp. 479–-493. Springer (2003)

\bibitem{ref:pp-05} Nourolhoda Alemi Neissi, Masoud Mazloom: GLS Optimization Algorithm for Solving Travelling Salesman Problem. In: Second Int. Conf. on Computer and Electrical Engineering, vol. 1, pp. 291–-294. IEEE Press (2009)

\bibitem{ref:pp-06} Bernd, F., Peter, M.: New Genetic Local Search Operators Traveling Salesman Problem. In: The 4th Int. Conf. On Parallel Problem Solving from Nature, pp. 890–-899. Springer (1996)

\bibitem{ref:pp-07} Bernd Freisleben, Peter Merz: New Genetic Local Search for the TSP: New Results. In: Int. Conf. on Evolutionary Computation, pp. 159-–164. IEEE Press (1997)

\bibitem{ref:pp-08} Freisleben, B., Merz, P.: A Genetic Local Search Algorithm for Solving Symmetric and Asymmetric Traveling Salesman Problems. In: Int. Conf. on Evolutionary Computation,  pp. 616–-621. IEEE Press (1996)

\bibitem{ref:url-09} TSPLIB, \url{http://comopt.ifi.uni-heidelberg.de/software/TSPLIB95/}

\bibitem{ref:pp-10} Renders, J.M.,Bersini, H.: Hybridizing genetic algorithms with hill-climbing methods for global optimization: two possible ways. In: IEEE World Congress on Computational Intelligence, vol. 1, pp. 312-–317. IEEE Press (1994)

\bibitem{ref:pp-11} Wan-rong Jih, Hsu, J.Y.-J.: Dynamic vehicle routing using hybrid genetic algorithms. In: Int. Conf. on Robotics \& Automation, vol. 1, pp. 453-–458. IEEE Press (1999)

\bibitem{ref:pp-12} Ray, S.S., Bandyopadhyay, S., Pal, S.K.: New operators of genetic algorithms for traveling salesman problem, ICPR-04, vol. 2, pp. 497-–500. Cambridge, UK, (2004)

\bibitem{ref:pp-13} Larranaga, P., Kuijpers, C., Murga, R., Inza, I., Dizdarevic, S.: Genetic algorithms for the traveling salesman problem: A review of representations and operators. In: Artificial Intelligence, vol. 13, pp. 129-–170. Kluwer Academic Publishers (1999) 

\bibitem{ref:pp-14} Held, M., Karp, R.M.: A dynamic programming approach to sequencing problems. In: Journal of the Society for Industrial and Applied Mathematics. vol. 10, pp. 196–-210 (1962)

\bibitem{ref:pp-15} Haykin, S.: Neural Networks: A Comprehensive Foundation, 2nd Edition. Prentice-Hall (1999)

\bibitem{ref:pp-16} Yagiura, M., Ibaraki, T.: The Use of Dynamic Programming in Genetic Algorithms for Permutation Problems. In: European Journal of Operational Research, vol. 92, pp. 387–-401 (1996)

\bibitem{ref:pp-17} Murat, A., Novruz A.: Development a new mutation operator to solve the Traveling Salesman Problem by aid of Genetic Algorithms. In: Expert Systems with Applications. vol. 38, pp. 1313-–1320. ScienceDirect (2011)


\bibitem{ref:pp-18} Olaf, M,, Bernd, B., Jakob, B., Heike, T., Markus, W., Frank, N.:  Local Search and the Traveling Salesman Problem: A Feature-Based Characterization of Problem Hardness. Lecture Notes in Computer Science, pp. 115-–129, Springer (2012)

\bibitem{ref:pp-19} Sourav, S., Anwesha, D., Satrughna, S.: Solution of traveling salesman problem on scx based selection with performance analysis using Genetic Algorithm. In: International Journal of Engineering Science and Technology (IJEST), vol. 3, pp. 6622-–6629 (2011)

\bibitem{ref:pp-20} Sehrawat, M., Singh, S.: Modified Order Crossover (OX) Operator. International Journal on Computer Science \&  Engineering, vol. 3, pp. 2019–-2014 (2011)

\bibitem{ref:pp-21} Shubhra, S.R., Sanghamitra, B., Sankar K.P.:  New Genetic Operators for Solving TSP: Application to Microarray Gene Ordering. PReMI, vol. 3776, pp. 617--622, Springer (2005)

\bibitem{ref:pp-22} Sallabi, O.M., El-Haddad, Y.: An Improved Genetic Algorithm to Solve the Traveling Salesman Problem. Proceedings of World Academy of Science: Engineering \& Technology, vol. 52, pp. 530--533 (2009)

\bibitem{ref:pp-23} Kusum, D., Hadush, M.: New Variations of Order Crossover for Travelling Salesman Problem. In: Int. Journal of Combinatorial Optimization Problems and Informatics, vol. 2, pp. 2--13 (2011)

\bibitem{ref:pp-24} Zakir, H.A.: Genetic Algorithm for the Traveling Salesman Problem using Sequential Constructive Crossover Operator. In: Int. Journal of Biometric and Bioinformatics, vol. 3, pp. 96--106 (2010)

\bibitem{ref:pp-25} Dr.Sabry M. Abdel-Moetty, Asmaa O. Heakil: Enhanced Traveling Salesman Problem Solving using Genetic Algorithm Technique with modified Sequential Constructive Crossover Operator. In: Int. Journal of Computer Science and Network Security. vol. 12, pp. 134--139 (2012)
\end{thebibliography}
\end{document}